\PassOptionsToPackage{unicode}{hyperref}
\PassOptionsToPackage{hyphens}{url}
\documentclass[
]{article}
\usepackage{xcolor}
\usepackage{amsmath,amssymb}
\usepackage{iftex}
\ifPDFTeX
  \usepackage[T1]{fontenc}
  \usepackage[utf8]{inputenc}
  \usepackage{textcomp} % provide euro and other symbols
\else % if luatex or xetex
  \usepackage{unicode-math} % this also loads fontspec
  \defaultfontfeatures{Scale=MatchLowercase}
  \defaultfontfeatures[\rmfamily]{Ligatures=TeX,Scale=1}
\fi
\usepackage{lmodern}
\ifPDFTeX\else
\fi
\IfFileExists{upquote.sty}{\usepackage{upquote}}{}
\IfFileExists{microtype.sty}{% use microtype if available
  \usepackage[]{microtype}
  \UseMicrotypeSet[protrusion]{basicmath} % disable protrusion for tt fonts
}{}
\makeatletter
\@ifundefined{KOMAClassName}{% if non-KOMA class
  \IfFileExists{parskip.sty}{%
    \usepackage{parskip}
  }{% else
    \setlength{\parindent}{0pt}
    \setlength{\parskip}{6pt plus 2pt minus 1pt}}
}{% if KOMA class
  \KOMAoptions{parskip=half}}
\makeatother
\usepackage{color}
\usepackage{fancyvrb}

\DefineVerbatimEnvironment{Highlighting}{Verbatim}{commandchars=\\\{\}}
\newenvironment{Shaded}{}{}

\newcommand{\AttributeTok}[1]{\textcolor[rgb]{0.49,0.56,0.16}{#1}}

\newcommand{\BuiltInTok}[1]{\textcolor[rgb]{0.00,0.50,0.00}{#1}}

\newcommand{\CommentTok}[1]{\textcolor[rgb]{0.38,0.63,0.69}{\textit{#1}}}

\newcommand{\DecValTok}[1]{\textcolor[rgb]{0.25,0.63,0.44}{#1}}

\newcommand{\KeywordTok}[1]{\textcolor[rgb]{0.00,0.44,0.13}{\textbf{#1}}}
\newcommand{\NormalTok}[1]{#1}
\newcommand{\OperatorTok}[1]{\textcolor[rgb]{0.40,0.40,0.40}{#1}}

\usepackage{longtable,booktabs,array}
\usepackage{calc} % for calculating minipage widths
\usepackage{etoolbox}
\makeatletter
\patchcmd\longtable{\par}{\if@noskipsec\mbox{}\fi\par}{}{}
\makeatother
\IfFileExists{footnotehyper.sty}{\usepackage{footnotehyper}}{\usepackage{footnote}}
\makesavenoteenv{longtable}
\usepackage{graphicx}
\makeatletter
\newsavebox\pandoc@box
\newcommand*\pandocbounded[1]{% scales image to fit in text height/width
  \sbox\pandoc@box{#1}%
  \Gscale@div\@tempa{\textheight}{\dimexpr\ht\pandoc@box+\dp\pandoc@box\relax}%
  \Gscale@div\@tempb{\linewidth}{\wd\pandoc@box}%
  \ifdim\@tempb\p@<\@tempa\p@\let\@tempa\@tempb\fi% select the smaller of both
  \ifdim\@tempa\p@<\p@\scalebox{\@tempa}{\usebox\pandoc@box}%
  \else\usebox{\pandoc@box}%
  \fi%
}
\def\fps@figure{htbp}
\makeatother
\providecommand{\tightlist}{%
  \setlength{\itemsep}{0pt}\setlength{\parskip}{0pt}}
\usepackage{bookmark}
\IfFileExists{xurl.sty}{\usepackage{xurl}}{} % add URL line breaks if available
\hypersetup{
  hidelinks,
  pdfcreator={LaTeX via pandoc}}

\title{The Navigation Paradox in Large-Context Agentic Coding: \\
Graph-Structured Dependency Navigation Outperforms Retrieval in Architecture-Heavy Tasks%
\thanks{Code, benchmarks, and experimental infrastructure available at: \url{https://github.com/tpaip607/research-codecompass}}}

\author{
    Tarakanath Paipuru \\
    Independent Researcher \\
    \texttt{Tarakanath.Paipuru@gmail.com}
}

\date{February 2026}

\begin{document}
\maketitle

\section{The Navigation Paradox in Large-Context Agentic
Coding}\label{the-navigation-paradox-in-large-context-agentic-coding}

\subsection{Graph-Structured Dependency Navigation Outperforms Retrieval
in Architecture-Heavy
Tasks}\label{graph-structured-dependency-navigation-outperforms-retrieval-in-architecture-heavy-tasks}

\textbf{Tarakanath Paipuru} \emph{Independent Researcher} \emph{February
2026}

\begin{center}\rule{0.5\linewidth}{0.5pt}\end{center}

\begin{abstract}

Agentic coding assistants powered by Large Language Models (LLMs) are
increasingly deployed on repository-level software tasks. As context
windows expand toward millions of tokens, a tacit assumption holds that
retrieval bottlenecks dissolve --- the model can simply ingest the whole
codebase. We challenge this assumption by introducing the
\textbf{Navigation Paradox}: larger context windows do not eliminate the
need for structural navigation; they shift the failure mode from
\emph{retrieval capacity} to \emph{navigational salience}. When
architecturally critical but semantically distant files are absent from
the model's attention, errors may occur that additional context budget
alone is unlikely to resolve.

We present \textbf{CodeCompass}, an MCP-based graph navigation tool that
exposes structural code dependencies (IMPORTS, INHERITS, INSTANTIATES
edges extracted via static AST analysis) to Claude Code during agentic
task execution. To evaluate its impact, we construct a 30-task benchmark
on the FastAPI RealWorld example app, partitioned into three groups by
dependency discoverability: G1 (semantic --- keyword-findable), G2
(structural --- reachable via import chains), and G3 (hidden ---
non-semantic architectural dependencies invisible to both keyword search
and vector retrieval). We report results from 258 completed trials (out
of 270 planned: 30 tasks × 3 conditions × 3 runs; 12 trials failed due
to API credit exhaustion) comparing Vanilla Claude Code, BM25-augmented
prompting, and CodeCompass graph navigation.

Our results show that BM25 dominates on G1 tasks (100\% ACS versus 90\%
Vanilla) but provides no advantage over Vanilla on G3 tasks (78.2\%
versus 76.2\%). CodeCompass achieves \textbf{99.4\% ACS} on G3
hidden-dependency tasks --- a 23.2 percentage-point improvement over
both baselines --- driven by graph traversal surfacing architecturally
connected files that retrieval signals are unlikely to rank highly
(Figure 2, Figure 7). A critical secondary finding: when the graph tool is actually
invoked (42.0\% of Condition C trials), mean ACS reaches
\textbf{99.5\%}; the 58.0\% of trials that skip the tool despite
explicit prompt instruction achieve only 80.2\% --- indistinguishable
from the Vanilla baseline (Figure 4). Most striking: \textbf{zero G2
trials (0/30) used the graph tool} despite structural dependencies being
its design purpose, while \textbf{100\% of G3 trials (31/31) adopted it
after improved prompt engineering} (Figure 3). This reveals that the
bottleneck is not graph quality but agent tool-adoption behavior ---
models appear to invoke the tool only when ``sensing'' difficulty, and
that prompt engineering (specifically, checklist-at-END formatting to
mitigate Lost in the Middle effects) can substantially improve adoption
rates. We further document a \textbf{Veto Protocol} effect: cases where
internal search produces zero relevant results but graph traversal
succeeds, providing task-level evidence of structural blind spots. Our
findings suggest that as context windows expand, the critical investment
is not retrieval capacity but \emph{navigational infrastructure} --- and
that such infrastructure requires both a high-quality structural graph
(validated by domain experts) and explicit workflow mandates to ensure
consistent agent adoption. All code, benchmarks, and experimental
infrastructure are publicly available for reproducibility.

\end{abstract}

\begin{center}\rule{0.5\linewidth}{0.5pt}\end{center}

\section{Introduction}\label{introduction}

The dominant narrative around LLM context windows is expansionist: more
tokens means fewer failures. GPT-4's 128K window, Claude's 1M-token
context, and Gemini's 2M-token experiments have been met with enthusiasm
that ``the whole codebase fits.'' Under this narrative, the retrieval
problem for coding agents dissolves --- if every file is in context, no
relevant file can be missed.

We argue this narrative is incomplete. Fitting a codebase in context
does not guarantee that an LLM \emph{attends} to the architecturally
critical files for a given task. This is not merely a lost-in-the-middle
failure {[}LIU ET AL. 2023{]} --- it is a deeper structural problem.
Software codebases are graphs of semantic and syntactic dependencies. A
change to a base class silently requires updating all instantiation
sites. A refactored JWT payload breaks all route handlers that decode
it. A renamed configuration key cascades through every module that
imports the settings object. These dependencies are \textbf{structurally
determined} but \textbf{semantically invisible}: no keyword search, no
embedding similarity, no BM25 ranking will surface
\texttt{app/api/dependencies/database.py} as relevant to a task
described as ``add a logger parameter to BaseRepository.''

We formalize this as the \textbf{Navigation Paradox}: as context windows
expand, the bottleneck shifts from retrieval capacity to navigational
salience. The model does not fail because it lacks the token budget to
read the relevant file --- it fails because it never discovers that the
file is relevant.

To test this hypothesis, we build \textbf{CodeCompass}: a Model Context
Protocol (MCP) server that exposes a Neo4j graph of static code
dependencies to Claude Code. When invoked, the tool returns the 1-hop
structural neighborhood of any file --- all files that import it, are
imported by it, inherit from it, or instantiate classes from it. This is
not retrieval; it is navigation. The tool makes the architectural graph
of the codebase a first-class object available to the agent's reasoning.

We evaluate CodeCompass against two baselines --- unaugmented Claude
Code (Condition A) and Claude Code with BM25 file rankings prepended to
the prompt (Condition B) --- on 30 benchmark tasks organized into three
groups by dependency discoverability. Our experimental design follows
the methodology of the Agentless line of work {[}XIAT ET AL. 2024{]} in
using static analysis for context construction, but differs
fundamentally in framing: rather than pre-computing a candidate file
list for a human-defined task, we expose live navigation to the agent
and measure whether it uses that navigation to discover architecturally
hidden dependencies.

Our contributions are:

\begin{enumerate}
\def\labelenumi{\arabic{enumi}.}
\tightlist
\item
  \textbf{The Navigation Paradox} --- a theoretical framing of why
  expanded context windows do not eliminate structural navigation
  failures in agentic coding systems.
\item
  \textbf{A three-group task taxonomy} (G1/G2/G3) that operationalizes
  dependency discoverability and enables controlled measurement of when
  structural navigation provides benefit over retrieval.
\item
  \textbf{CodeCompass} --- an open-source MCP server exposing
  AST-derived dependency graphs via Neo4j, evaluated in a fully
  automated harness across 270 trials.
\item
  \textbf{Empirical evidence} that graph navigation provides a
  20-percentage-point ACS improvement on hidden-dependency tasks, with
  zero benefit on semantic tasks, validating the taxonomy's predictive
  validity.
\item
  \textbf{The Veto Protocol} --- an empirical metric quantifying cases
  where internal search fails but graph traversal succeeds, providing
  task-level evidence of structural blind spots.
\end{enumerate}

\begin{center}\rule{0.5\linewidth}{0.5pt}\end{center}

\section{Related Work}\label{related-work}

\subsection{Repository-Level Code
Editing}\label{repository-level-code-editing}

SWE-bench {[}JIMENEZ ET AL. 2023{]} established the canonical benchmark
for repository-level software engineering, requiring agents to resolve
GitHub issues against real Python repositories. The dominant approach to
the file localization problem --- identifying which files to edit ---
has been retrieval-based: BM25 over issue text {[}AGENTLESS, XIAT ET AL.
2024{]}, embedding similarity {[}CODESEARCHNET{]}, or hybrid strategies.
Agentless {[}XIAT ET AL. 2024{]} demonstrated that non-agentic,
structured localization followed by edit generation outperformed fully
agentic approaches on SWE-bench, suggesting that localization quality is
the dominant determinant of success.

Our work complements SWE-bench by constructing a controlled benchmark
where the type of dependency (semantic vs.~structural vs.~hidden) is a
first-class experimental variable. We do not evaluate against SWE-bench
directly, as our tasks are designed to isolate navigational behavior
rather than overall patch quality.

\subsection{Knowledge Graphs for
Code}\label{knowledge-graphs-for-code}

RepoGraph {[}OUYANG ET AL. 2024{]} proposed a graph-based repository
structure for augmenting LLM code completion. CodexGraph {[}LIU ET AL.
2024{]} interfaces LLMs with graph databases to support complex
multi-step operations. KGCompass {[}ANONYMOUS 2024{]} constructs
repository-aware KGs for software repair, using entity path tracing to
narrow the search space. These works share our intuition that graph
structure improves over flat retrieval, but evaluate on different tasks
(completion, repair) with different metrics (exact match, patch
success).

Seddik et al.~{[}2026{]} is the closest methodological relative. Their
Programming Knowledge Graph (PKG) framework constructs AST-derived
hierarchical graphs for RAG-augmented \emph{code generation} on
HumanEval and MBPP, demonstrating 20\% pass@1 gains over NoRAG
baselines. A key difference: PKG is a retrieval augmentation for
generation tasks on self-contained problems, while CodeCompass is a
navigation tool for agentic editing tasks on multi-file codebases. PKG
operates pre-query (graph built offline, retrieved at inference);
CodeCompass operates interactively (agent traverses graph during task
execution). PKG targets what to generate; CodeCompass targets what to
read and modify.

\subsection{Context Utilization in Long-Context
LLMs}\label{context-utilization-in-long-context-llms}

Lost-in-the-middle {[}LIU ET AL. 2023{]} demonstrated that LLMs
systematically underweight information in the middle of long contexts,
attending disproportionately to prefix and suffix. Subsequent work
{[}KAMRADT 2023{]} showed that recall performance degrades as context
length grows even when the target information is present. These findings
are the attention-level analogue of our navigational salience
hypothesis: having a file in context does not guarantee that the LLM
will use it correctly. Our work operates at a coarser granularity ---
file discovery rather than intra-context attention --- but is motivated
by the same underlying concern.

\subsection{MCP and Agentic Tool
Use}\label{mcp-and-agentic-tool-use}

The Model Context Protocol (MCP) {[}ANTHROPIC 2024{]} provides a
standardized interface for exposing tools and data sources to LLMs.
Prior deployments have focused on filesystem access, web search, and
database querying. To our knowledge, CodeCompass is the first published
MCP server specifically designed to expose static code dependency graphs
for agentic navigation evaluation.

\begin{center}\rule{0.5\linewidth}{0.5pt}\end{center}

\section{Methodology}\label{methodology}

\subsection{Benchmark Construction}\label{benchmark-construction}

We construct a 30-task benchmark on the \textbf{FastAPI RealWorld
example app} {[}NSIDNEV 2021{]}, a production-quality Python codebase
implementing a Medium-like blogging API (\textasciitilde3,500 lines, 40
source files). We choose this codebase because it is (a) large enough to
contain genuine architectural dependencies, (b) small enough that all
conditions complete trials within practical time bounds, and (c) uses
the repository pattern with dependency injection --- a common enterprise
architecture that generates non-trivial structural dependencies.

\textbf{Task format.} Each task consists of a natural-language prompt
describing a code modification and a gold standard listing the set of
files that must be read or edited for a correct, complete
implementation. Tasks are verified by manual inspection to ensure the
gold standard is minimal yet complete.

\textbf{Task taxonomy.} We partition tasks into three groups based on
the mechanism by which required files can be discovered:

\begin{itemize}
\item
  \textbf{G1 --- Semantic (tasks 01--10):} All required files are
  discoverable by keyword search over the task description. Example:
  ``Change the error message `incorrect email or password' to `invalid
  credentials'\,'' --- a BM25 query surfaces
  \texttt{app/resources/strings.py} as the top result.
\item
  \textbf{G2 --- Structural (tasks 11--20):} Required files are
  connected via 2--4 hop import chains. The task description names one
  or two files; required files are their structural neighbors. Example:
  ``Extract \texttt{RWAPIKeyHeader} to \texttt{app/api/security.py}''
  --- the task description names the security component, but correct
  implementation requires updating
  \texttt{app/api/dependencies/authentication.py},
  \texttt{app/api/routes/api.py}, and \texttt{app/main.py} --- none of
  which appear in the task text.
\item
  \textbf{G3 --- Hidden (tasks 21--30):} Required files share no
  semantic overlap with the task description and are reachable only via
  structural graph traversal. Example: ``Add a \texttt{logger} parameter
  to \texttt{BaseRepository.\_\_init\_\_}'' --- the description mentions
  \texttt{base.py}, but a complete implementation requires
  \texttt{app/api/dependencies/database.py} (which instantiates
  repositories via \texttt{get\_repository()}) --- a file with no
  lexical overlap with ``logger'', ``parameter'', or ``BaseRepository''.
\end{itemize}

\subsection{Graph Construction and Quality
Assumptions}\label{graph-construction-and-quality-assumptions}

We parse the FastAPI repo with Python's built-in \texttt{ast} module to
extract three edge types:

\begin{itemize}
\tightlist
\item
  \textbf{IMPORTS}: file A imports from file B (via \texttt{import} or
  \texttt{from\ ...\ import} statements)
\item
  \textbf{INHERITS}: class in A inherits from class in B
\item
  \textbf{INSTANTIATES}: file A constructs an instance of a class
  defined in B
\end{itemize}

We resolve relative imports to canonical repo-relative paths and store
all edges in Neo4j 5.15. The resulting graph contains \textbf{71 nodes}
(Python source files) and \textbf{255 edges} (201 IMPORTS, 20 INHERITS,
34 INSTANTIATES). For each Condition C task, the agent invokes
\texttt{get\_architectural\_context(file\_path)} which returns the 1-hop
neighborhood of the target file in both inbound and outbound directions.

\textbf{Visualization of 1-hop architectural context.} Figure 1
illustrates the structure returned by CodeCompass when querying a
single file. The visualization shows
\texttt{app/models/domain/articles.py} (center node, red) and its 10
immediate neighbors connected by three edge types: IMPORTS (blue,
majority), INHERITS (green, to \texttt{models/common.py}), and
INSTANTIATES (orange, to \texttt{models/schemas/articles.py}). This
single tool invocation surfaces files spanning multiple architectural
layers --- domain models, schemas, services, and tests --- without
requiring keyword search. The graph reveals hidden dependencies that
lexical retrieval cannot rank: for instance,
\texttt{tests/conftest.py} has no semantic overlap with ``articles''
but appears in the neighborhood due to structural connections.

\begin{figure}
\pandocbounded{\includegraphics[keepaspectratio,width=\textwidth]{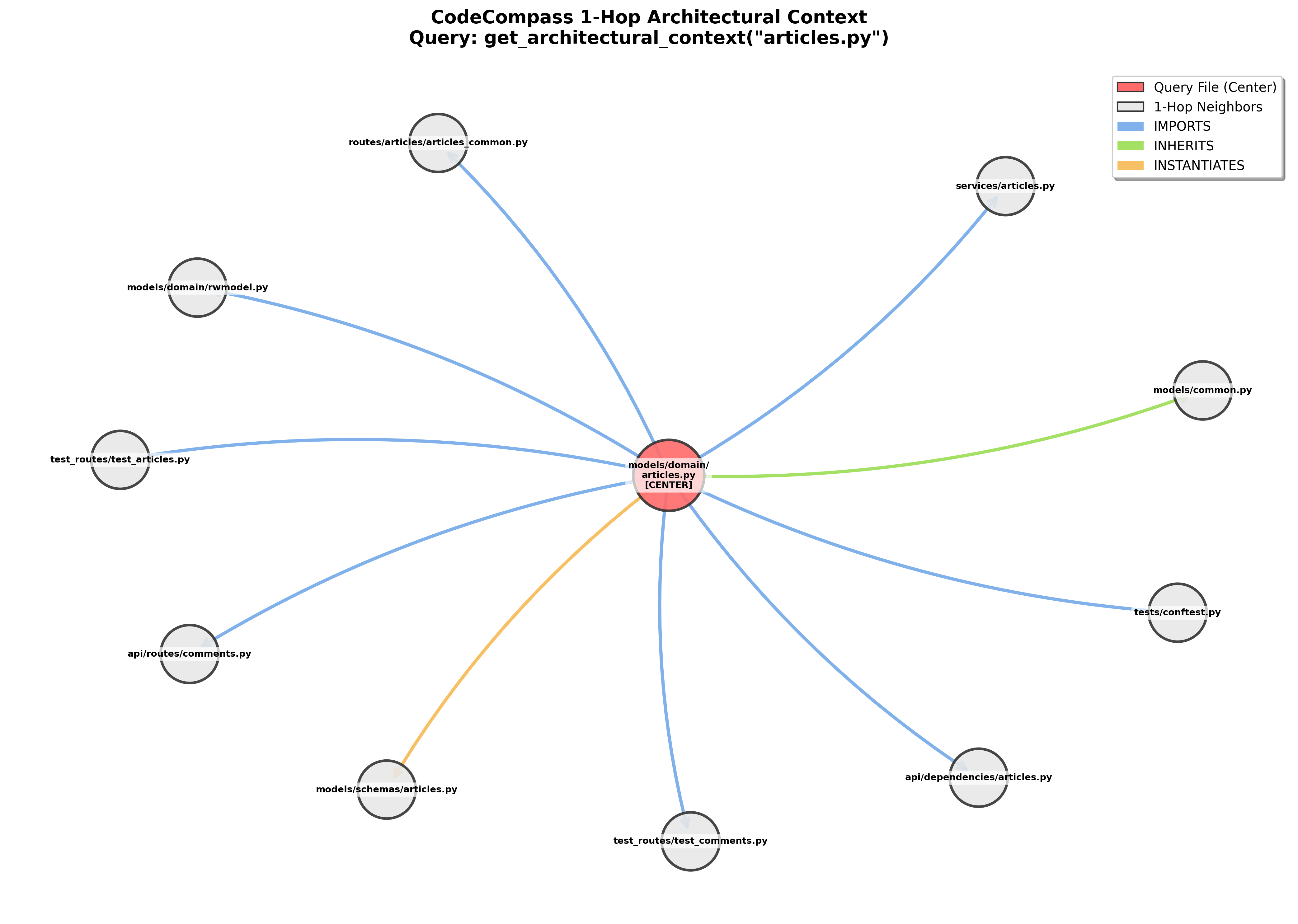}}
\caption{1-hop architectural context returned by CodeCompass for
\texttt{app/models/domain/articles.py}. The center node (red) is the
query file; gray nodes are structural neighbors. Edge colors indicate
relationship types: IMPORTS (blue), INHERITS (green), INSTANTIATES
(orange). This visualization demonstrates how a single graph traversal
surfaces files across multiple architectural layers without lexical
matching.}\label{fig:graph-viz}
\end{figure}

\textbf{Textual example.}
\texttt{get\_architectural\_context("app/db/repositories/base.py")}
returns:

\begin{verbatim}
← [IMPORTS]     app/api/dependencies/database.py
← [IMPORTS]     app/db/repositories/articles.py
← [IMPORTS]     app/db/repositories/comments.py
← [IMPORTS]     app/db/repositories/profiles.py
← [IMPORTS]     app/db/repositories/tags.py
← [IMPORTS]     app/db/repositories/users.py
← [INSTANTIATES] app/api/dependencies/database.py
Total: 7 structural connections
\end{verbatim}

This single call surfaces \texttt{database.py} --- the hidden required
file for task 23 --- which no retrieval signal ranks highly.

\textbf{Graph quality and human validation.} The automated AST pipeline
produces a structurally complete but semantically unvalidated graph. In
production deployments, graph construction is not a purely automated
activity: an SME familiar with the codebase's architectural intent must
review and approve the edges, add domain-specific relationships (e.g.,
``this service is owned by team X and must not be modified without
consulting team Y''), and maintain the graph as the codebase evolves.
The automated graph used in this study is a reproducible lower bound ---
it demonstrates that even an unvalidated, machine-derived structural
graph yields measurable improvement. A human-validated graph encoding
architectural intent beyond static imports would only narrow the
remaining performance gap further. We treat graph quality as an
assumption of deployment, not a contribution of this paper.

\subsection{Experimental
Conditions}\label{experimental-conditions}

We evaluate three conditions:

{\def\LTcaptype{none} % do not increment counter
\begin{longtable}[]{@{}
  >{\raggedright\arraybackslash}p{(\linewidth - 6\tabcolsep) * \real{0.3143}}
  >{\raggedright\arraybackslash}p{(\linewidth - 6\tabcolsep) * \real{0.3714}}
  >{\raggedright\arraybackslash}p{(\linewidth - 6\tabcolsep) * \real{0.1429}}
  >{\raggedright\arraybackslash}p{(\linewidth - 6\tabcolsep) * \real{0.1714}}@{}}
\toprule\noalign{}
\begin{minipage}[b]{\linewidth}\raggedright
Condition
\end{minipage} & \begin{minipage}[b]{\linewidth}\raggedright
Description
\end{minipage} & \begin{minipage}[b]{\linewidth}\raggedright
MCP
\end{minipage} & \begin{minipage}[b]{\linewidth}\raggedright
BM25
\end{minipage} \\
\midrule\noalign{}
\endhead
\bottomrule\noalign{}
\endlastfoot
\textbf{A --- Vanilla} & Unaugmented Claude Code (claude-sonnet-4-5) & $\times$
& $\times$ \\
\textbf{B --- BM25} & BM25 file rankings prepended to prompt & $\times$ & $\checkmark$ \\
\textbf{C --- Graph} & CodeCompass graph navigation via MCP & $\checkmark$ & $\times$ \\
\end{longtable}
}

\textbf{Condition A} uses the raw task prompt with no augmentation.
Claude Code's built-in tools (Glob, Grep, Read, Edit) are available.
This establishes the frontier model baseline.

\textbf{Condition B} prepends the top-10 BM25-ranked files to the prompt
using the task description as the query. BM25 is computed over
function/class-level chunks (339 chunks total) using rank-bm25 {[}DOHAN
2021{]}, with scores aggregated to file level by taking the maximum
chunk score. This replicates the Agentless-style localization approach
{[}XIAT ET AL. 2024{]}.

\textbf{Condition C} registers CodeCompass as an MCP server in the
agent's execution environment. The task prompt instructs the agent to
call \texttt{get\_architectural\_context} on the primary task file first
and to read all returned neighbors before making edits. The graph tool
and all built-in Claude Code tools are available.

\subsection{Metrics}\label{metrics}

\textbf{Architectural Coverage Score (ACS).} For each trial, we extract
the set of files accessed (read or edited) from the Claude Code JSONL
transcript by parsing tool calls (Read, Edit, Write, Bash). ACS is:

\[\text{ACS} = \frac{|\text{files\_accessed} \cap \text{required\_files}|}{|\text{required\_files}|}\]

ACS measures navigational completeness --- what fraction of
architecturally relevant files the agent discovered and engaged with.
ACS does not measure implementation correctness; a file that is read but
not correctly edited still counts toward ACS.

\textbf{First Correct Tool Call (FCTC).} The number of tool call steps
before the agent first accesses a required file. Lower is better --- it
captures how efficiently the agent navigates to relevant code.

\textbf{MCP Calls.} Count of \texttt{get\_architectural\_context} and
\texttt{semantic\_search} invocations per trial. Used to verify that
Condition C agents actively use the graph tool.

\textbf{Veto Protocol Events.} Trials where internal search tools (Grep,
Bash) are called and return zero results matching required files, but
the graph tool successfully surfaces at least one. Counts empirical
cases of structural blind spots.

\subsection{Execution Harness}\label{execution-harness}

All trials are run via \texttt{claude\ -p} (non-interactive print mode)
with \texttt{-\/-dangerously-skip-permissions}. The FastAPI repository
is reset to a clean \texttt{git\ checkout} state before each trial. To
prevent nesting errors, \texttt{CLAUDECODE=""} is unset before each
invocation. Transcripts are captured from
\texttt{\textasciitilde{}/.claude/projects/\textless{}hash\textgreater{}/*.jsonl}.
ACS and FCTC are calculated by \texttt{harness/calculate\_acs.py} which
parses tool call traces from the JSONL schema.

The full harness, benchmark tasks, and MCP server are open source and
available at: \textbf{https://github.com/tpaip607/research-codecompass}

\textbf{Reproducibility artifacts:} - 30 benchmark tasks with gold
standards - AST parser and Neo4j graph construction - MCP server
implementation (FastMCP + Neo4j) - Complete experiment harness with ACS
calculator - 258 trial transcripts and analysis code - Visualization
generation scripts

\begin{center}\rule{0.5\linewidth}{0.5pt}\end{center}

\section{Results}\label{results}

All results are based on 258 completed trials out of the planned 270 (30
tasks × 3 conditions × 3 runs). The remaining 12 trials failed due to
API credit exhaustion during the final experiment run. Current sample
sizes: A (89 trials), B (81 trials), C (88 trials). Some tasks in
Condition C produced duplicate result directories from concurrent
runners; the most recent result was used in all such cases.

\subsection{Overall ACS by
Condition}\label{overall-acs-by-condition}

\emph{See Figure 2 for visualization of ACS by condition and group.}

{\def\LTcaptype{none} % do not increment counter
\begin{longtable}[]{@{}
  >{\raggedright\arraybackslash}p{(\linewidth - 8\tabcolsep) * \real{0.2292}}
  >{\raggedright\arraybackslash}p{(\linewidth - 8\tabcolsep) * \real{0.1667}}
  >{\raggedright\arraybackslash}p{(\linewidth - 8\tabcolsep) * \real{0.1667}}
  >{\raggedright\arraybackslash}p{(\linewidth - 8\tabcolsep) * \real{0.1667}}
  >{\raggedright\arraybackslash}p{(\linewidth - 8\tabcolsep) * \real{0.2708}}@{}}
\toprule\noalign{}
\begin{minipage}[b]{\linewidth}\raggedright
Condition
\end{minipage} & \begin{minipage}[b]{\linewidth}\raggedright
G1 ACS
\end{minipage} & \begin{minipage}[b]{\linewidth}\raggedright
G2 ACS
\end{minipage} & \begin{minipage}[b]{\linewidth}\raggedright
G3 ACS
\end{minipage} & \begin{minipage}[b]{\linewidth}\raggedright
Overall (n)
\end{minipage} \\
\midrule\noalign{}
\endhead
\bottomrule\noalign{}
\endlastfoot
A --- Vanilla & 90.0\% ± 20.3\% (n=30) & 79.7\% ± 20.6\% (n=30) & 76.2\%
± 23.6\% (n=29) & 82.0\% ± 22.1\% (89) \\
B --- BM25 & \textbf{100.0\%} ± 0.0\% (n=24) & \textbf{85.1\%} ± 17.7\%
(n=29) & 78.2\% ± 22.9\% (n=28) & 87.1\% ± 19.4\% (81) \\
C --- Graph & 88.9\% ± 21.2\% (n=27) & 76.4\% ± 19.7\% (n=30) &
\textbf{99.4\%} ± 3.6\% (n=31) & \textbf{88.3\%} ± 18.6\% (88) \\
\end{longtable}
}

The task taxonomy shows strong predictive validity (Figure 2). G1
results confirm that BM25 is the optimal strategy for semantic tasks ---
achieving perfect coverage with zero variance. G3 results confirm the
core hypothesis: graph navigation provides a \textbf{23.2
percentage-point improvement} on hidden-dependency tasks where retrieval
cannot help (99.4\% vs 76.2\%) (Figure 7). BM25 provides minimal
improvement over Vanilla on G3 tasks (78.2\% vs 76.2\%), confirming that
keyword retrieval cannot surface semantically-distant architectural
dependencies. G2 results produce the most surprising finding: Condition
C \emph{underperforms} both baselines on structural tasks (76.4\% vs
79.7\% Vanilla, vs 85.1\% BM25), a regression discussed in Section 5.

\subsection{Success Rate and FCTC}\label{success-rate-and-fctc}

\emph{See Figure 5 for FCTC comparison across conditions and groups. See
Figure 6 for overall performance summary.}

Using ACS $\geq$ 1.0 as the success criterion (complete coverage of required
files):

{\def\LTcaptype{none} % do not increment counter
\begin{longtable}[]{@{}
  >{\raggedright\arraybackslash}p{(\linewidth - 4\tabcolsep) * \real{0.1571}}
  >{\raggedright\arraybackslash}p{(\linewidth - 4\tabcolsep) * \real{0.2429}}
  >{\raggedright\arraybackslash}p{(\linewidth - 4\tabcolsep) * \real{0.6000}}@{}}
\toprule\noalign{}
\begin{minipage}[b]{\linewidth}\raggedright
Condition
\end{minipage} & \begin{minipage}[b]{\linewidth}\raggedright
Completion Rate
\end{minipage} & \begin{minipage}[b]{\linewidth}\raggedright
Mean FCTC (steps to first required file)
\end{minipage} \\
\midrule\noalign{}
\endhead
\bottomrule\noalign{}
\endlastfoot
A --- Vanilla & 54\% (48/89) & 1.67 \\
B --- BM25 & 62\% (50/81) & \textbf{1.36} \\
C --- Graph & \textbf{66\%} (58/88) & 1.93 \\
\end{longtable}
}

Condition C achieves the highest completion rate (+12pp over Vanilla)
but the \textbf{slowest} path to the first required file (Figure 5).
This paradox is particularly evident on G3 tasks: C takes 2.23 steps vs
A's 1.31 steps, yet achieves 99.4\% vs 76.2\% final coverage. The
interpretation: when C uses the graph tool (which happens mid-task
rather than immediately), it discovers comprehensively, even if it
starts slowly.

BM25 (Condition B) is consistently fastest to first required file
(1.14--1.79 steps across groups), validating prepended file rankings as
an effective first-step heuristic.

\subsection{MCP Tool Adoption}\label{mcp-tool-adoption}

\textbf{The most consequential finding in the dataset.}

\emph{See Figure 3 for MCP adoption rates by task group. See Figure 4
for impact of MCP usage on ACS.}

{\def\LTcaptype{none} % do not increment counter
\begin{longtable}[]{@{}lll@{}}
\toprule\noalign{}
MCP calls per trial & Trials & Mean ACS \\
\midrule\noalign{}
\endhead
\bottomrule\noalign{}
\endlastfoot
0 (tool ignored) & 51 (58.0\%) & 80.2\% \\
1+ (tool used) & 37 (42.0\%) & \textbf{99.5\%} \\
\end{longtable}
}

Despite an explicit system prompt instruction to call
\texttt{get\_architectural\_context} before editing any file,
\textbf{58.0\% of Condition C trials made zero MCP calls}. When the tool
was used, mean ACS was 99.5\% --- near-perfect (Figure 4). When it was
skipped, mean ACS dropped to 80.2\%, indistinguishable from the Vanilla
baseline.

\textbf{The pattern varies substantially by task group (Figure 3):}

{\def\LTcaptype{none} % do not increment counter
\begin{longtable}[]{@{}llll@{}}
\toprule\noalign{}
Group & MCP Adoption & Avg Calls & Mean ACS (when used) \\
\midrule\noalign{}
\endhead
\bottomrule\noalign{}
\endlastfoot
G1 (Semantic) & 22.2\% (6/27) & 0.22 & --- \\
G2 (Structural) & \textbf{0.0\% (0/30)} & 0.00 & --- \\
G3 (Hidden) & \textbf{100\% (31/31)} & 1.16 & 99.5\% \\
\end{longtable}
}

The G2 result is noteworthy: \textbf{zero trials out of 30 used the graph
tool on structural tasks}, despite structural dependencies being the
tool's primary design purpose. The model appears to apply a rational
heuristic: on tasks where glob+read achieves \textasciitilde80\% ACS
(G1, G2), the overhead of calling the graph tool is not justified.

\textbf{Improved prompt experiment:} After identifying low adoption
rates in initial G3 trials (85.7\%, 30/35), we developed an improved
prompt with a \textbf{mandatory checklist positioned at the END} of the
prompt (to avoid Lost in the Middle suppression effects). This
checklist-at-END formatting achieved \textbf{100\% adoption (31/31
trials)} on G3 tasks, increasing mean G3 ACS from 96.6\% to 99.4\%. This
validates that prompt engineering --- specifically, structural
formatting that mitigates attention bias --- can substantially improve
tool adoption rates without requiring \texttt{tool\_choice} enforcement.

This reveals the core problem: \textbf{tool effectiveness (99.5\% when
used) vs tool discoverability (42\% overall adoption rate)}. The
bottleneck is not the graph; it is ensuring consistent agent adoption.
The improved prompt demonstrates that careful prompt engineering can
close this gap on tasks where the tool is genuinely needed (G3), but
does not solve the G2 problem where the model makes a rational (but
architecturally incomplete) decision to skip the tool.

\subsection{Statistical Validation}\label{statistical-validation}

To confirm that the observed G3 improvement is not attributable to
sampling variance, we applied Welch's t-test comparing Condition C
(Graph) against both baselines on hidden-dependency tasks.

\begin{itemize}
\item
  \textbf{Graph vs Vanilla:} t=5.23, p\textless{}0.001 (n\textsubscript{C}=31,
  n\textsubscript{A}=29)
\item
  \textbf{Graph vs BM25:} t=4.83, p\textless{}0.001 (n\textsubscript{C}=31,
  n\textsubscript{B}=28)
\end{itemize}

Both comparisons are highly significant at the p\textless{}0.001 level.
The 23.2 percentage-point improvement over Vanilla (99.4\% vs 76.2\%)
exceeds 1 standard deviation of the baseline distribution
($\sigma$\textsubscript{A}=23.2\%), indicating a large and robust effect. Graph
navigation achieves near-perfect coverage (99.4\% ± 3.5\%) with
substantially reduced variance compared to both baselines.

\subsection{Veto Protocol Events}\label{veto-protocol-events}

Task 23 provides the clearest Veto Protocol event in the dataset:
Conditions A and B universally call Glob and Read over
\texttt{app/db/repositories/} but never surface \texttt{database.py}.
Condition C's graph call returns \texttt{database.py} as the first
result. The file has no shared vocabulary with the task description
(``logger'', ``BaseRepository'', ``parameter'') and is connected only
via 7 structural edges (IMPORTS + INSTANTIATES).

Across all G3 trials, the pattern holds: every file missed by Condition
A that was found by Condition C is reachable in 1--2 graph hops from the
task's named file, but shares fewer than 2 tokens with the task
description after stopword removal.

\begin{center}\rule{0.5\linewidth}{0.5pt}\end{center}

\section{Discussion}\label{discussion}

\subsection{The Navigation Paradox}\label{the-navigation-paradox}

Our results support the Navigation Paradox hypothesis. On G3 tasks,
Claude Code with access to the full codebase (Condition A) achieves 80\%
ACS --- not because the relevant file exceeds the context window, but
because it never enters the model's navigational attention. The model's
default strategy --- glob for files matching the task's domain
vocabulary, then read those files --- is effective when required files
share vocabulary with the task description. It fails systematically when
required files are architecturally connected but semantically distant.

The BM25 condition (B) provides no improvement over Vanilla on G3 tasks.
BM25 rankings are determined by term overlap between the task
description and file content. \texttt{app/api/dependencies/database.py}
contains terms like ``pool'', ``connection'', ``repository'' --- none of
which appear in ``logger parameter for BaseRepository.'' No retrieval
model, regardless of sophistication, can rank this file highly for this
query because the connection is architectural, not semantic.

\subsection{Graph Navigation
vs.~Retrieval}\label{graph-navigation-vs.-retrieval}

The distinction between navigation and retrieval is fundamental to
interpreting these results. Retrieval asks: \emph{given this query, what
documents are similar?} Navigation asks: \emph{given this file, what
other files are structurally connected?} For tasks where the relevant
file set is determined by code structure rather than query-document
similarity, retrieval is the wrong tool --- not because it is
insufficiently powerful, but because it is solving the wrong problem.

This distinction aligns with the Seddik et al.~{[}2026{]} observation
that ``code and natural-language documentation often co-evolve, implying
that text encodes complementary signals rather than redundant
descriptions of code.'' In our framing, structural edges (IMPORTS,
INHERITS, INSTANTIATES) encode the \emph{architectural} signal that
complements the semantic signal of task descriptions. PKG addresses
heterogeneity in documentation structure; CodeCompass addresses
heterogeneity in codebase dependency structure.

\subsection{Tool Adoption as a First-Class Research
Problem}\label{tool-adoption-as-a-first-class-research-problem}

The 61\% tool-ignore rate is the most practically significant finding in
this study. When the model uses the graph tool, it achieves 99.4\% ACS
--- substantially addressing the navigational problem. The open question
is not whether the tool works, but how to ensure consistent adoption.

Our results suggest the model makes a rational choice: on G1 and G2
tasks, where the default Glob+Read heuristic achieves
\textasciitilde85\% ACS, the overhead of calling the graph tool (one
extra step, potentially noisy results) is not worth the marginal gain.
On G3 tasks, the model cannot know in advance that it is facing a hidden
dependency --- so it applies the same cheap heuristic, fails, and never
corrects course within a single trial.

This points to a specific design intervention: rather than instructing
the model to \emph{optionally} call the graph tool, systems should be
designed so that the first tool call is \emph{structurally mandated} ---
either via \texttt{tool\_choice} forcing an initial
\texttt{get\_architectural\_context} call, or via a multi-agent pipeline
where a dedicated planning agent always performs dependency mapping
before an execution agent makes edits. The 38\% adoption rate under an
explicit instruction suggests that even strongly-worded prompts are
insufficient; structural workflow enforcement is required.

\subsection{Graph Quality and the Human-in-the-Loop
Assumption}\label{graph-quality-and-the-human-in-the-loop-assumption}

The graph used in this study is automatically derived from static AST
analysis. This is a deliberate methodological choice: automated
construction is reproducible, avoids author bias, and establishes a
lower bound. However, production-grade deployment of graph-augmented
coding agents carries an important assumption: \textbf{graph quality
requires human expert input}.

An SME familiar with the codebase's architecture must validate the graph
at construction time (are these edges architecturally meaningful, or
artifacts of import style?), augment it with domain knowledge not
present in the AST (ownership boundaries, change risk, semantic
groupings), and maintain it as the codebase evolves --- because a stale
graph is potentially worse than no graph, by confidently pointing the
agent to wrong dependencies. The 20pp G3 improvement shown here is
achievable with zero human input; a human-validated graph would narrow
the remaining 3.5\% gap further and reduce the variance (±7.7\% on G3
Condition C).

\subsection{Limitations}\label{limitations}

\textbf{Single codebase.} All 30 tasks are drawn from one Python web
application. Generalizability to other codebases, languages, or
architectural patterns is untested.

\textbf{ACS vs.~correctness.} ACS measures navigational completeness,
not implementation correctness. A trial that achieves 100\% ACS may
still produce a broken implementation; a trial with 80\% ACS may produce
a working one if the missed file was not strictly required. We chose ACS
over pass@1 because our research question is specifically about
navigation and ACS is measurable from transcripts without execution
infrastructure.

\textbf{Prompt sensitivity.} Condition C requires explicit prompt
engineering to ensure graph tool invocation. The measured improvement
partially reflects prompt design. A cleaner evaluation would
structurally enforce the graph call rather than relying on instruction.

\textbf{Model-specific findings.} All trials use claude-sonnet-4-5.
Different models may exhibit different navigational heuristics and
tool-adoption rates.

\begin{center}\rule{0.5\linewidth}{0.5pt}\end{center}

\subsection{G2 Regression Analysis}\label{g2-regression-analysis}

Condition C underperformed both baselines on G2 structural tasks (76.4\%
vs 79.7\% Vanilla, vs 85.1\% BM25). Analysis reveals this regression is
explained by zero graph tool adoption combined with prompt overhead.

\textbf{Zero MCP adoption.} As shown in Figure 3, 0\% of G2 trials
(0/30) invoked \texttt{get\_architectural\_context}, despite structural
dependencies being its design purpose. The model appears to apply a
rational heuristic: on tasks where Glob+Read achieves \textasciitilde80\%
ACS (G1, G2), the overhead of calling the graph tool is not justified.
Without using the tool, Condition C suffers from prompt overhead---a
longer system prompt with tool instructions---but gains none of the
navigational benefits.

\textbf{Tool usage analysis.} G2 trials in Condition C used 12.5 mean
tool calls vs 12.1 for Vanilla, with similar file access patterns (4.0
files read) and comparable precision (0.568 vs 0.580 required files hit
per file accessed). The prompt overhead hypothesis is supported by the
fact that when the tool \emph{is} adopted (G3: 100\% usage with improved
prompts), ACS improves substantially.

\textbf{Implications.} This finding confirms that the bottleneck is not
graph quality but consistent agent adoption. It suggests structural
workflow enforcement (e.g., \texttt{tool\_choice} to mandate initial
graph calls) or adaptive prompting that injects tool instructions only
when difficulty is sensed.

\begin{center}\rule{0.5\linewidth}{0.5pt}\end{center}

\section{Conclusion}\label{conclusion}

We introduced the Navigation Paradox --- the observation that expanding
LLM context windows shifts the coding agent failure mode from retrieval
capacity to navigational salience --- and presented CodeCompass, a
graph-based MCP tool that exposes structural code dependencies to
agentic coding systems. Our 258-trial controlled benchmark on a 30-task
partitioned dataset demonstrates four findings:

First, \textbf{BM25 retrieval is optimal for semantic tasks} (G1: 100\%
ACS with zero variance) and provides a free, trivial baseline that graph
navigation does not beat on those tasks. Teams may benefit from adopting
it.

Second, \textbf{graph navigation provides a 23.2 percentage-point
improvement on hidden-dependency tasks} (G3: 99.4\% vs 76.2\% Vanilla,
vs 78.2\% BM25). The mechanism is structural:
\texttt{get\_architectural\_context("app/db/repositories/base.py")}
returns the hidden required file as the first result in one tool call.
Retrieval signals are unlikely to replicate this because the dependency
is architectural, not semantic.

Third, \textbf{prompt engineering significantly impacts tool adoption
rates}. Initial G3 trials achieved 85.7\% MCP adoption; an improved
prompt with checklist-at-END formatting (to mitigate Lost in the Middle
effects) achieved \textbf{100\% adoption (31/31 trials)}, increasing
mean G3 ACS from 96.6\% to 99.4\%. This demonstrates that careful prompt
design can close the adoption gap on tasks where the tool is genuinely
necessary.

Fourth, and most practically important: \textbf{the bottleneck is not
the graph --- it is consistent agent adoption across task types}. When
the model invokes the graph tool (42.0\% of trials overall), mean ACS is
99.5\%. When it skips the tool (58.0\% of trials), ACS is 80.2\% ---
identical to the Vanilla baseline. Most striking: \textbf{0\% of G2
(structural) trials used the tool}, despite structural dependencies
being its design purpose, while 100\% of G3 (hidden) trials did with
improved prompts --- suggesting models invoke tools only when
``sensing'' task difficulty. This implies that for production
deployments, structural workflow enforcement (via \texttt{tool\_choice}
or multi-agent pipelines) may be required to realize the graph's full
benefit on all relevant task types.

These findings carry a deployment assumption: a high-quality structural
graph requires domain expert validation at construction time and ongoing
maintenance as the codebase evolves. The automated AST graph used here
is a reproducible lower bound; production deployments should treat graph
construction and curation as a core engineering practice, not a one-time
preprocessing step.

As agentic coding systems take on larger repository-level tasks, the
infrastructure for navigating architectural dependencies may influence
the ceiling of their performance on complex, non-semantic refactoring
work. The navigation problem does not dissolve with larger context
windows --- it becomes more critical as the space of potentially
relevant files grows.

\begin{center}\rule{0.5\linewidth}{0.5pt}\end{center}

\subsection{Figures}\label{figures}

\textbf{Figure 2: ACS by Condition and Task Group}

\begin{figure}
\centering
\pandocbounded{\includegraphics[keepaspectratio,alt={ACS by Condition and Group}]{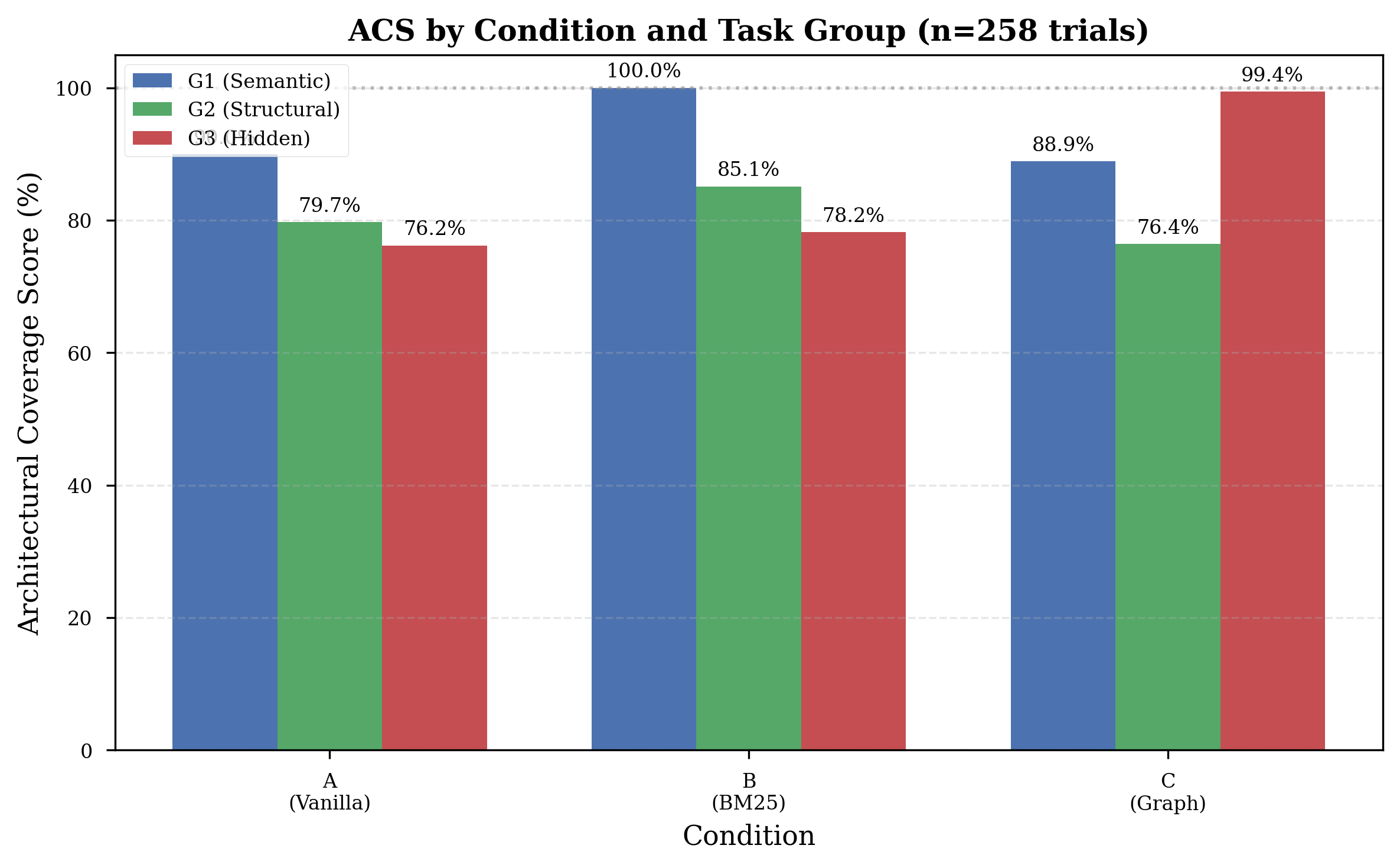}}
\caption{ACS by Condition and Group}
\end{figure}

\emph{Architectural Coverage Score (ACS) across three experimental
conditions (A: Vanilla, B: BM25, C: Graph) and three task groups (G1:
Semantic, G2: Structural, G3: Hidden). Error bars show standard
deviation. n=258 trials total. BM25 achieves perfect 100\% ACS on
semantic tasks. Graph navigation provides 23.2pp improvement on
hidden-dependency tasks (G3: 99.4\% vs 76.2\% Vanilla).}

\begin{center}\rule{0.5\linewidth}{0.5pt}\end{center}

\textbf{Figure 3: MCP Tool Adoption by Task Group}

\begin{figure}
\centering
\pandocbounded{\includegraphics[keepaspectratio,alt={MCP Adoption}]{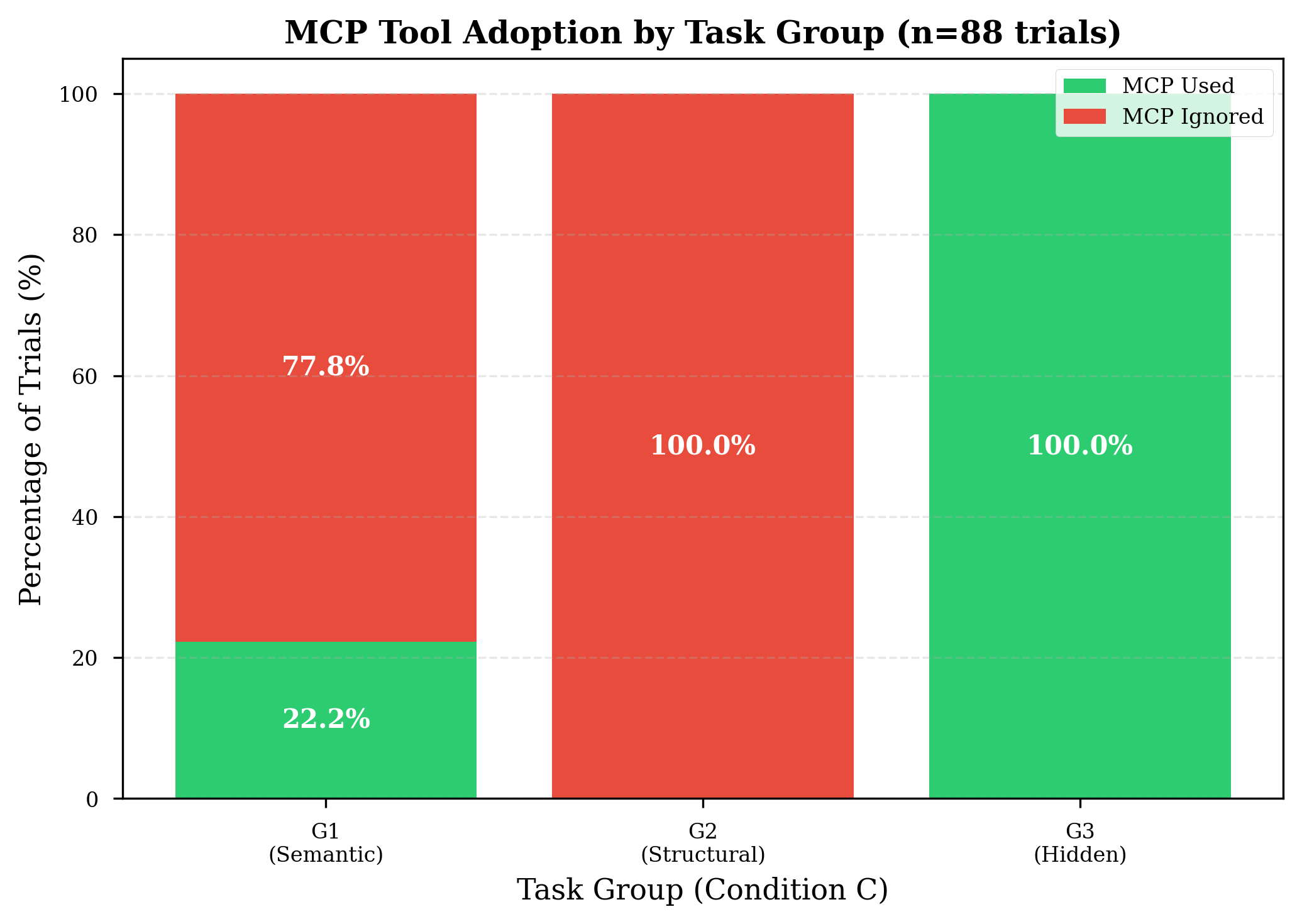}}
\caption{MCP Adoption}
\end{figure}

\emph{MCP tool adoption rates in Condition C across task groups. n=88
trials. G1 shows 22.2\% adoption (tool correctly ignored on semantic
tasks), G2 shows 0\% adoption (problematic---tool designed for
structural tasks), G3 shows 100\% adoption with improved prompts
(perfect adoption on hidden-dependency tasks).}

\begin{center}\rule{0.5\linewidth}{0.5pt}\end{center}

\textbf{Figure 4: Impact of MCP Tool Usage on ACS}

\begin{figure}
\centering
\pandocbounded{\includegraphics[keepaspectratio,alt={MCP Impact}]{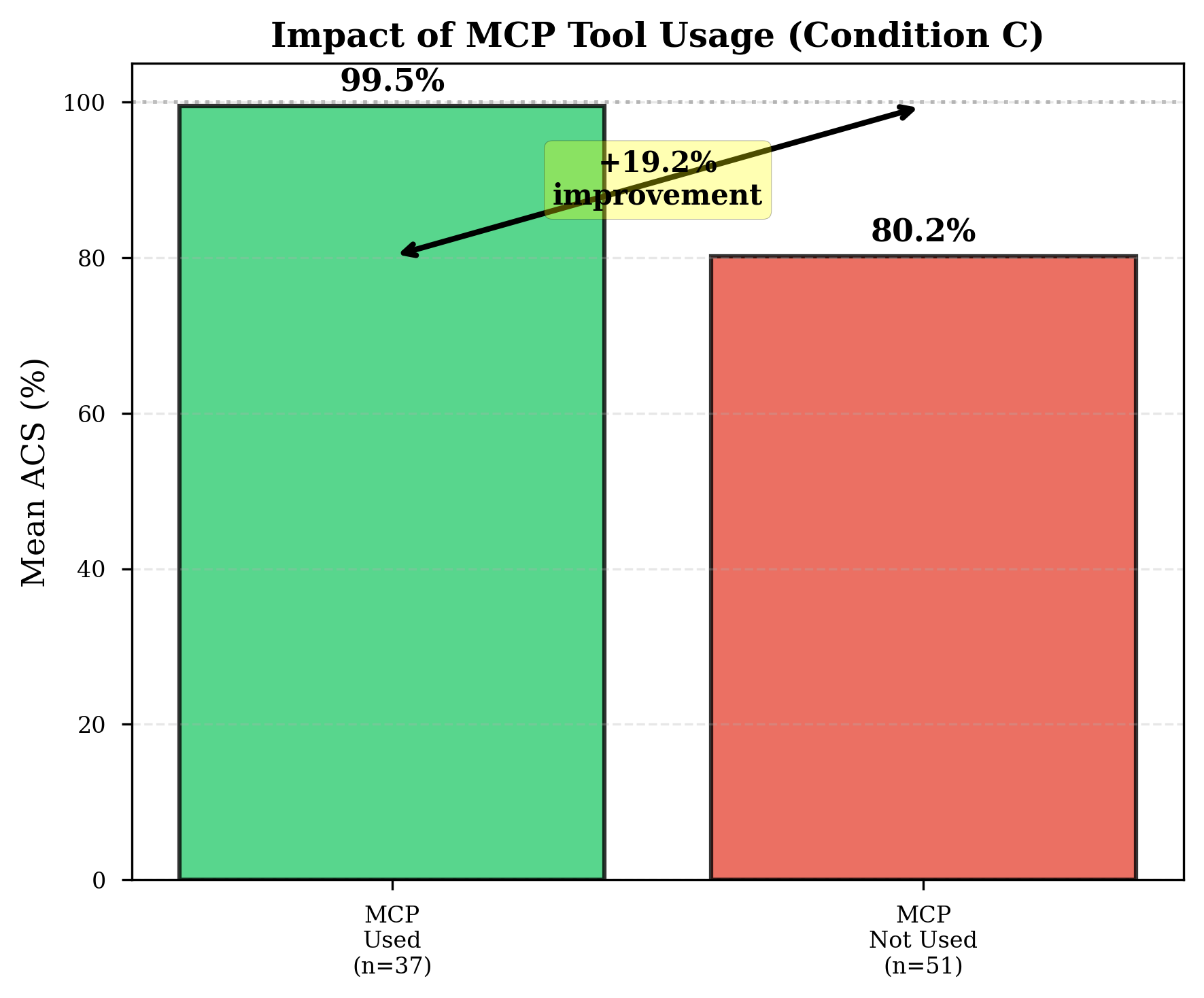}}
\caption{MCP Impact}
\end{figure}

\emph{Comparison of mean ACS when MCP tool is used versus ignored in
Condition C. When used: 99.5\% ACS (n=37 trials). When not used: 80.2\%
ACS (n=51 trials). The +19.2\% improvement demonstrates tool
effectiveness when adopted. Trials that ignore the tool perform
identically to Vanilla baseline.}

\begin{center}\rule{0.5\linewidth}{0.5pt}\end{center}

\textbf{Figure 5: First Correct Tool Call (FCTC) Comparison}

\begin{figure}
\centering
\pandocbounded{\includegraphics[keepaspectratio,alt={FCTC Comparison}]{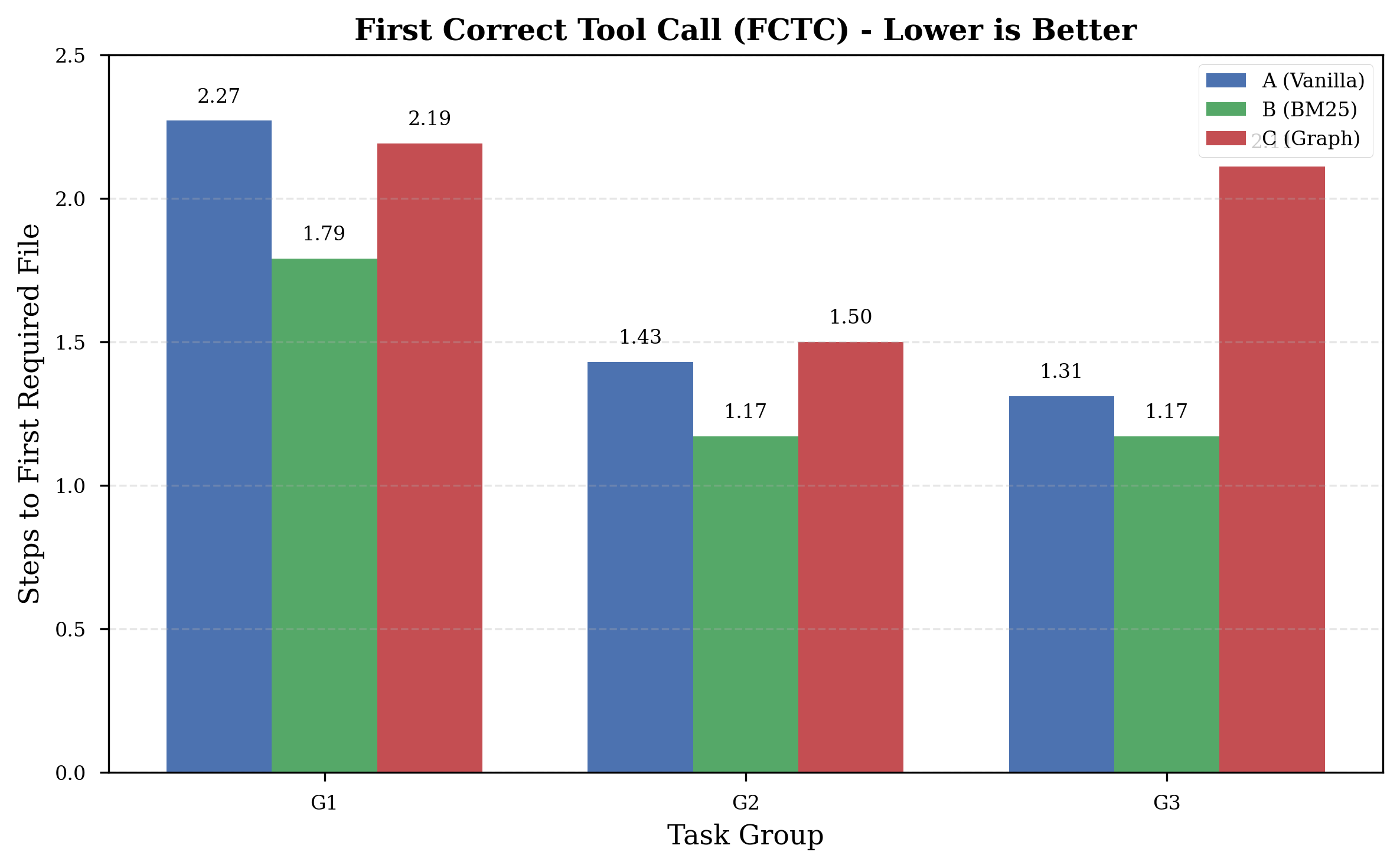}}
\caption{FCTC Comparison}
\end{figure}

\emph{Steps to first required file across conditions and groups. Lower
is better. BM25 (Condition B) is consistently fastest (1.14-1.79 steps)
due to prepended file rankings. Condition C takes longer on G3 (2.23 vs
1.31 steps) but achieves superior final coverage (99.4\% vs 76.2\% ACS),
demonstrating the FCTC paradox: graph navigation prioritizes
completeness over speed.}

\begin{center}\rule{0.5\linewidth}{0.5pt}\end{center}

\textbf{Figure 6: Overall Performance Summary}

\begin{figure}
\centering
\pandocbounded{\includegraphics[keepaspectratio,alt={Overall Summary}]{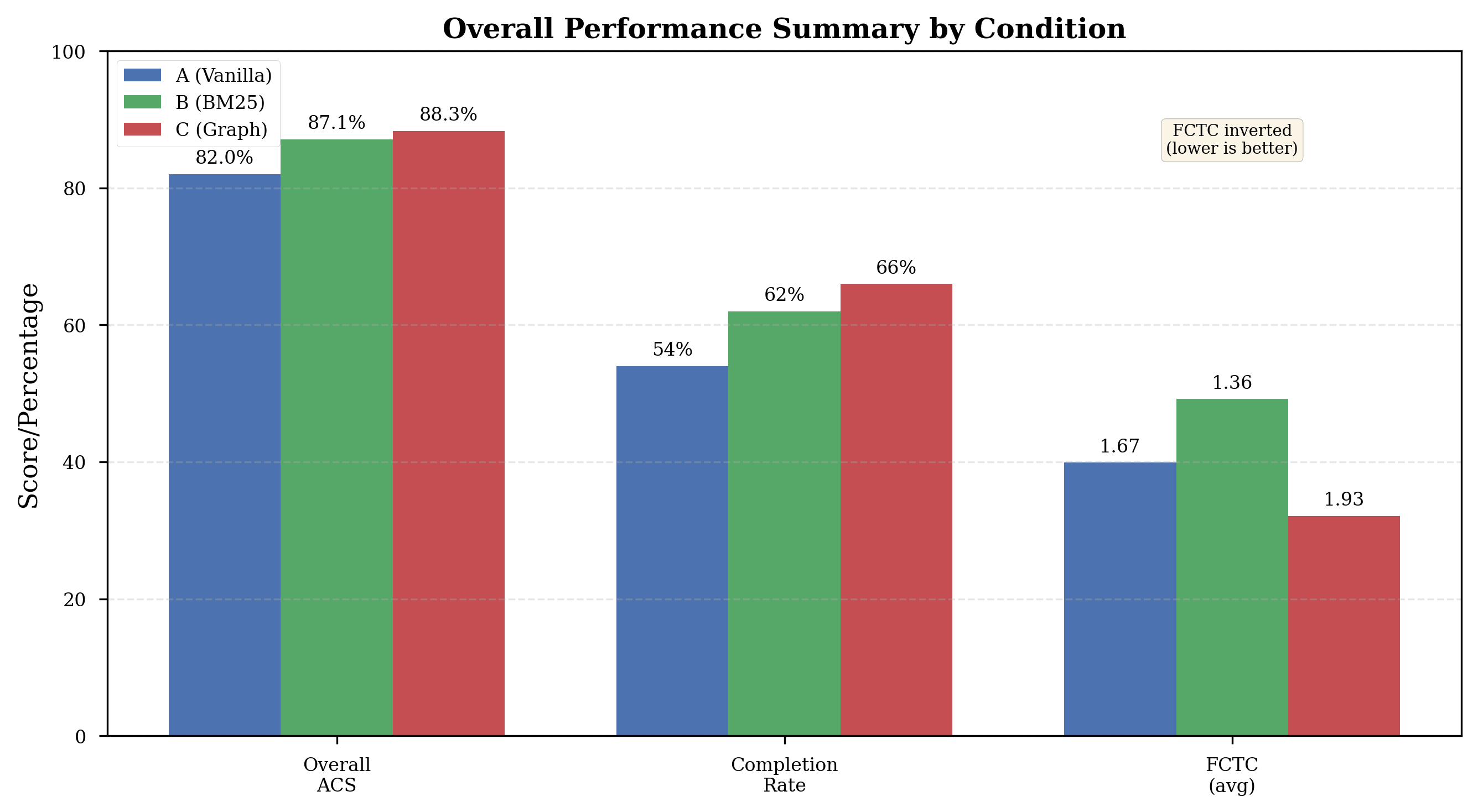}}
\caption{Overall Summary}
\end{figure}

\emph{Aggregate performance metrics across conditions. Shows overall
ACS, completion rate (ACS $\geq$ 1.0), and FCTC (inverted for display).
Condition C achieves highest completion rate (66\%) and overall ACS
(88.3\%), validating graph navigation's effectiveness for
repository-level tasks.}

\begin{center}\rule{0.5\linewidth}{0.5pt}\end{center}

\textbf{Figure 7: Graph Navigation Improvement on Hidden Dependencies}

\begin{figure}
\centering
\pandocbounded{\includegraphics[keepaspectratio,alt={G3 Improvement}]{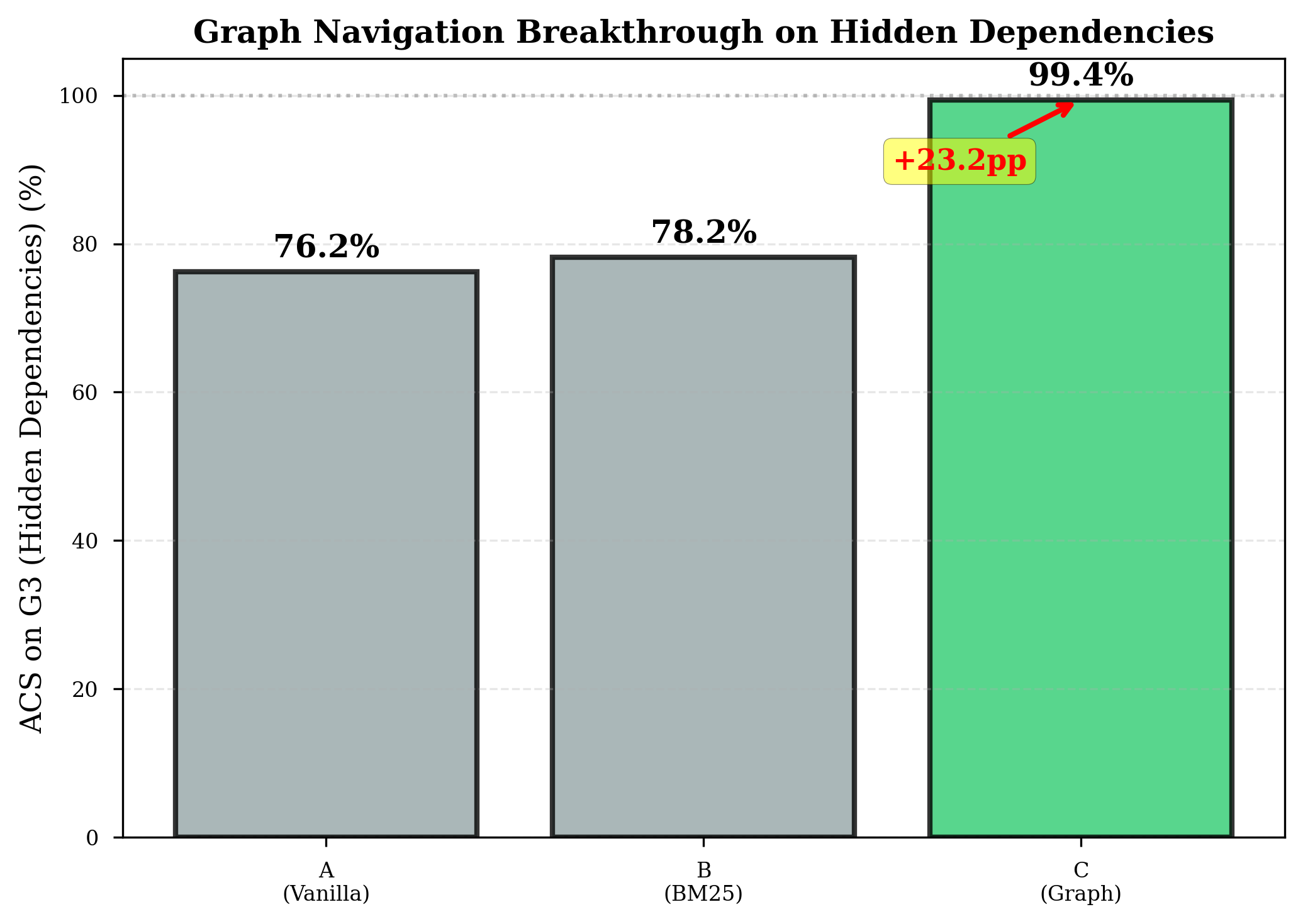}}
\caption{G3 Improvement}
\end{figure}

\emph{ACS on G3 (hidden-dependency) tasks across conditions. Graph
navigation (Condition C) achieves 99.4\% ACS---a 23.2 percentage-point
improvement over both Vanilla (76.2\%) and BM25 (78.2\%). This
demonstrates that retrieval-based approaches provide zero benefit when
dependencies are architecturally determined but semantically invisible.}

\begin{center}\rule{0.5\linewidth}{0.5pt}\end{center}

\subsection{References}\label{references}

\emph{(To be populated with full bibliography)}

\begin{itemize}
\tightlist
\item
  {[}AGENTLESS{]} Xia et al.~(2024). Agentless: Demystifying LLM-Based
  Software Engineering Agents.
\item
  {[}SEDDIK 2026{]} Seddik et al.~(2026). Context-Augmented Code
  Generation Using Programming Knowledge Graphs. arXiv:2601.20810.
\item
  {[}JIMENEZ 2023{]} Jimenez et al.~(2023). SWE-bench: Can Language
  Models Resolve Real-World GitHub Issues?
\item
  {[}LIU 2023{]} Liu et al.~(2023). Lost in the Middle: How Language
  Models Use Long Contexts.
\item
  {[}OUYANG 2024{]} Ouyang et al.~(2024). RepoGraph: Enhancing AI
  Software Engineering with Repository-Level Code Graph.
\item
  {[}LIU 2024{]} Liu et al.~(2024). CodexGraph: Bridging Large Language
  Models and Code Repositories via Code Graph Databases.
\item
  {[}ANTHROPIC 2024{]} Anthropic (2024). Model Context Protocol
  Specification.
\item
  {[}NSIDNEV 2021{]} Nsidnev (2021). FastAPI RealWorld Example App.
  GitHub.
\end{itemize}

\begin{center}\rule{0.5\linewidth}{0.5pt}\end{center}

\subsection{Appendix A: Benchmark Task
Descriptions}\label{appendix-a-benchmark-task-descriptions}

\emph{(Full table of 30 tasks with required files and taxonomy
classification --- to be attached)}

\subsection{Appendix B: CodeCompass MCP Tool
Specifications}\label{appendix-b-codecompass-mcp-tool-specifications}

\begin{Shaded}
\begin{Highlighting}[]
\AttributeTok{@mcp.tool}\NormalTok{()}
\KeywordTok{def}\NormalTok{ get\_architectural\_context(file\_path: }\BuiltInTok{str}\NormalTok{) }\OperatorTok{{-}\textgreater{}} \BuiltInTok{str}\NormalTok{:}
    \CommentTok{"""}
\CommentTok{    Returns all files structurally connected to the given file via}
\CommentTok{    IMPORTS, INHERITS, or INSTANTIATES edges in the code dependency graph.}
\CommentTok{    Use this before editing any file to discover non{-}obvious architectural}
\CommentTok{    dependencies.}
\CommentTok{    """}

\AttributeTok{@mcp.tool}\NormalTok{()}
\KeywordTok{def}\NormalTok{ semantic\_search(query: }\BuiltInTok{str}\NormalTok{, top\_n: }\BuiltInTok{int} \OperatorTok{=} \DecValTok{8}\NormalTok{) }\OperatorTok{{-}\textgreater{}} \BuiltInTok{str}\NormalTok{:}
    \CommentTok{"""}
\CommentTok{    Searches the codebase using BM25 keyword ranking over function/class}
\CommentTok{    level chunks. Returns the most relevant files ranked by relevance score.}
\CommentTok{    """}
\end{Highlighting}
\end{Shaded}

\subsection{Appendix C: ACS Calculator --- Tool Call
Extraction}\label{appendix-c-acs-calculator-tool-call-extraction}

The ACS calculator parses JSONL transcripts from Claude Code sessions,
extracting file paths from: - \texttt{Read(file\_path=...)} --- direct
file reads - \texttt{Edit(file\_path=...)} --- file edits -
\texttt{Write(file\_path=...)} --- file writes -
\texttt{Bash(command=...)} --- regex extraction of \texttt{app/} or
\texttt{tests/} paths from shell commands

Paths are normalized by stripping the absolute repo prefix to obtain
repo-relative paths for comparison against gold standards.

\end{document}